\documentclass[journal]{IEEEtran}
\usepackage{hyperref}
\usepackage{amsmath}
\usepackage{graphicx}		

\usepackage{svg}
\usepackage{academicons}
\newcommand{\orcidicon}{\includegraphics[width=8pt]{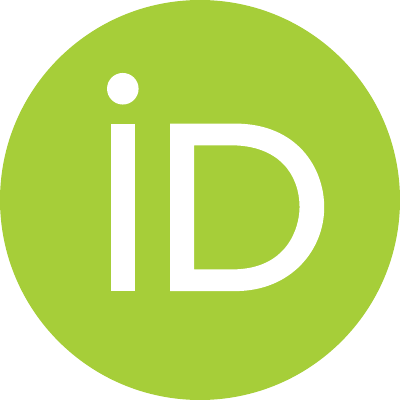}}

\usepackage{arydshln}
\usepackage[normalem]{ulem}
\usepackage[linesnumbered,ruled,vlined]{algorithm2e}
\usepackage{amssymb}
\usepackage{xcolor}
\usepackage{algpseudocode}
\usepackage{booktabs}
\usepackage{multirow}
\usepackage{graphicx}
\usepackage{colortbl}
\usepackage{makecell}
\usepackage{enumitem}
\usepackage{hyperref}
\usepackage{listings}
\usepackage[T1]{fontenc}
\usepackage[utf8]{inputenc}

\usepackage{bm}
\ifCLASSINFOpdf
\else
\fi
\begin{document}
%

\title{NBBOX: Noisy Bounding Box Improves \\Remote Sensing Object Detection \vspace{-0.2cm}}

%
%
%


\author{\IEEEauthorblockN{Yechan Kim\textsuperscript{\href{https://orcid.org/0000-0002-2438-3590}{\orcidicon{}}}, \textit{Student Member}, \textit{IEEE}}, 
\IEEEauthorblockN{SooYeon Kim\textsuperscript{\href{https://orcid.org/0009-0005-1474-6828}{\orcidicon{}}}, \textit{Student Member}, \textit{IEEE}}, and\\
\IEEEauthorblockN{Moongu Jeon\textsuperscript{\href{https://orcid.org/0000-0002-2775-7789}{\orcidicon{}}}, \textit{Senior Member}, \textit{IEEE}}
\vspace{-1cm}
\thanks{Received 14 September 2024; revised 16 December 2024; accepted 06 January 2025; Date of publication 00 00000000000 2025; date of current version 07 January 2025. This work was supported by the Agency For Defense Development Grant funded by the Korean Government (UI220066WD). (\textit{Corresponding authors}: \textit{Moongu Jeon and Yechan Kim.})}
\thanks{Yechan Kim, SooYeon Kim, and Moongu Jeon are with the School of Electrical Engineering and Computer Science, Gwangju Institute of Science and Technology (GIST), Gwangju 61005, South Korea (e-mail: 
    \href{mailto:yechankim@gm.gist.ac.kr}{yechankim@gm.gist.ac.kr}, 
    \href{mailto:bluesooyeon@gm.gist.ac.kr}{bluesooyeon@gm.gist.ac.kr},
    \href{mailto:mgjeon@gist.ac.kr}{mgjeon@gist.ac.kr}).}
\thanks{This article has supplementary downloadable material available at \url{https://ieeexplore.ieee.org/}, provided by the authors.}
\thanks{Digital Object Identifier xxxxxxxxxxxxxxxxxxxxxxxxxxxxx}}

%
%

\markboth{}%
{Shell \MakeLowercase{\textit{et al.}}: Bare Demo of IEEEtran.cls for Journals}
%



\maketitle

\begin{abstract}
Data augmentation has shown significant advancements in computer vision to improve model performance over the years, particularly in scenarios with limited and insufficient data.
    Currently, most studies focus on adjusting the image or its features to expand the size, quality, and variety of samples during training in various tasks including object detection.
    However, we argue that it is necessary to investigate bounding box transformations as a data augmentation technique rather than image-level transformations, especially in aerial imagery due to potentially inconsistent bounding box annotations.
    Hence, this letter presents a thorough investigation of bounding box transformation in terms of scaling, rotation, and translation for remote sensing object detection.
    We call this augmentation strategy \textbf{NBBOX} (\underline{N}oise Injection into \underline{B}ounding \underline{Box}).
    We conduct extensive experiments on DOTA and DIOR-R, both well-known datasets that include a variety of rotated generic objects in aerial images.
    Experimental results show that our approach significantly improves remote sensing object detection without whistles and bells and it is more time-efficient than other state-of-the-art augmentation strategies.
\end{abstract}


%
\IEEEpeerreviewmaketitle

\section{Introduction}
%
%
%
%
\IEEEPARstart{I}{n} recent years, remote sensing object detection has achieved remarkable success primarily due to the advancements in deep learning.
    Particularly, modern deep learning architectures such as convolutional neural networks and transformers have significantly enhanced the capabilities of object detection in Earth vision (e.g. \cite{dai2022tardet, li2023learning}).
    These models have millions or billions of parameters to learn, requiring tremendous training data to avoid the over-fitting problem.
    Compared to natural scenes (e.g. ImageNet \cite{deng2009imagenet} or MS-COCO \cite{lin2014microsoft}), datasets for overhead imagery usually lack diversity and quantity due to the expense of data collection. 
    To overcome this limitation, data augmentation plays a significant role as an implicit regularization strategy for model learning.

    
Existing visual data augmentation methods can be mainly divided into three categories: (a) image manipulation (e.g. \cite{devries2017improved, zhong2020random, yun2019cutmix, takahashi2019data}), (b) feature transformation (e.g. \cite{kuo2020featmatch, li2021feature}), and (c) generative image synthesis (e.g. \cite{goodfellow2014generative, zhu2017unpaired, trabucco2023effective}).
    For (a), geometric transformations (such as flipping and shearing), sharpness transformations, noise distribution, etc are considered the simplest augmentations.
    More advanced strategies in image manipulation contain image erasing and image mix.
    The image erasing like \cite{devries2017improved, zhong2020random} randomly selects a sub-region in an image and deletes its contents, while the image mix like \cite{yun2019cutmix, takahashi2019data} combines two or more images (or patches) as a single image during training.
    Meanwhile, (b) aims to directly manipulate features in the latent space rather than that raw input image. For example, \cite{kuo2020featmatch} applies augmentations to the extracted features and encourages the model to be consistent for both the original and augmented features. \cite{li2021feature} leverages the synergy between feature normalization and data augmentation to improve the robustness and accuracy of models.
    Unfortunately, both are hard to directly apply to rotated object detection.
    Finally, (c) produces samples with generative models based on GAN \cite{goodfellow2014generative} or Diffusion \cite{trabucco2023effective}. Such methods can generate high-quality (for GAN) or high-fidelity (for Diffusion) samples, but these require significant computational resources.  
    
While most prior work for data augmentation concentrates on adjusting the image or its features, only few have focused on the deformation of bounding boxes for object detection.
    However, \cite{murrugarra2022can} discovers the presence of low-confidence annotations in current overhead object detection datasets: there is a mismatch between the minimum enclosing box (i.e. optimal) and the actual annotation.
    Thus, we argue that investigating bounding box transformation is necessary to boost the model performance for remote sensing object detection, by enabling robust training in potentially inaccurate bounding boxes.
    
In this letter, we propose a simple but efficient data augmentation method named \textbf{NBBOX} (\underline{N}oise Injection into \underline{B}ounding \underline{Box}) for remote sensing object detection.
    Intuitively, our method adds noise to bounding boxes with geometric transformations during training (not for test/deployment).
    Particularly, this work thoroughly investigates bounding box transformation in terms of scaling, rotation, and translation.
    To demonstrate the effectiveness of our approach, extensive experiments are conducted on DOTA \cite{xia2018dota} and DIOR-R \cite{cheng2022anchor}, both datasets that contain rotated and densely-placed objects.
    
\begin{figure*}[ht!]
    \centering
    \includegraphics[width=16cm]{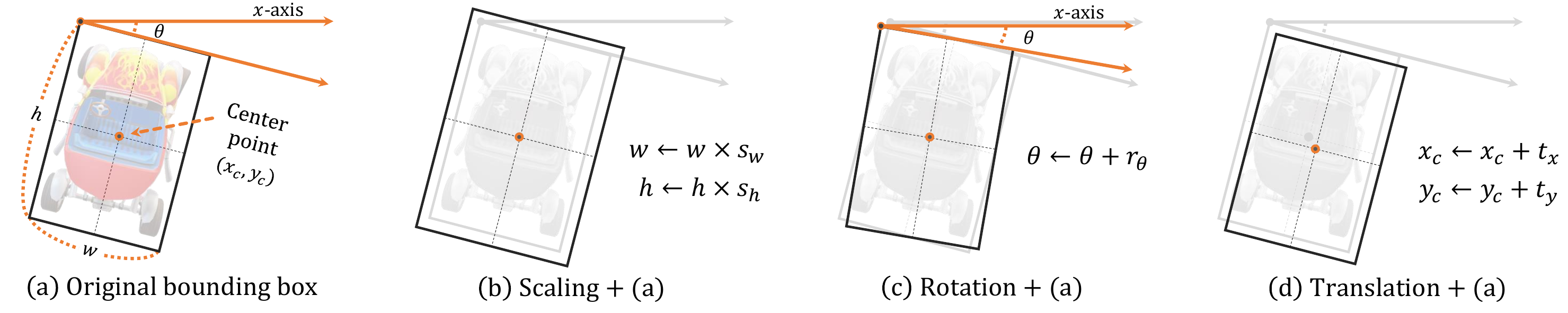}
    \caption{\small{Examples of the proposed data augmentation method named \textbf{NBBOX} for remote sensing object detection.}}
    \label{fig1}
   \vspace{-0.3cm}
\end{figure*}

\begin{figure}[ht]
    \centering
    \includegraphics[height=5cm]{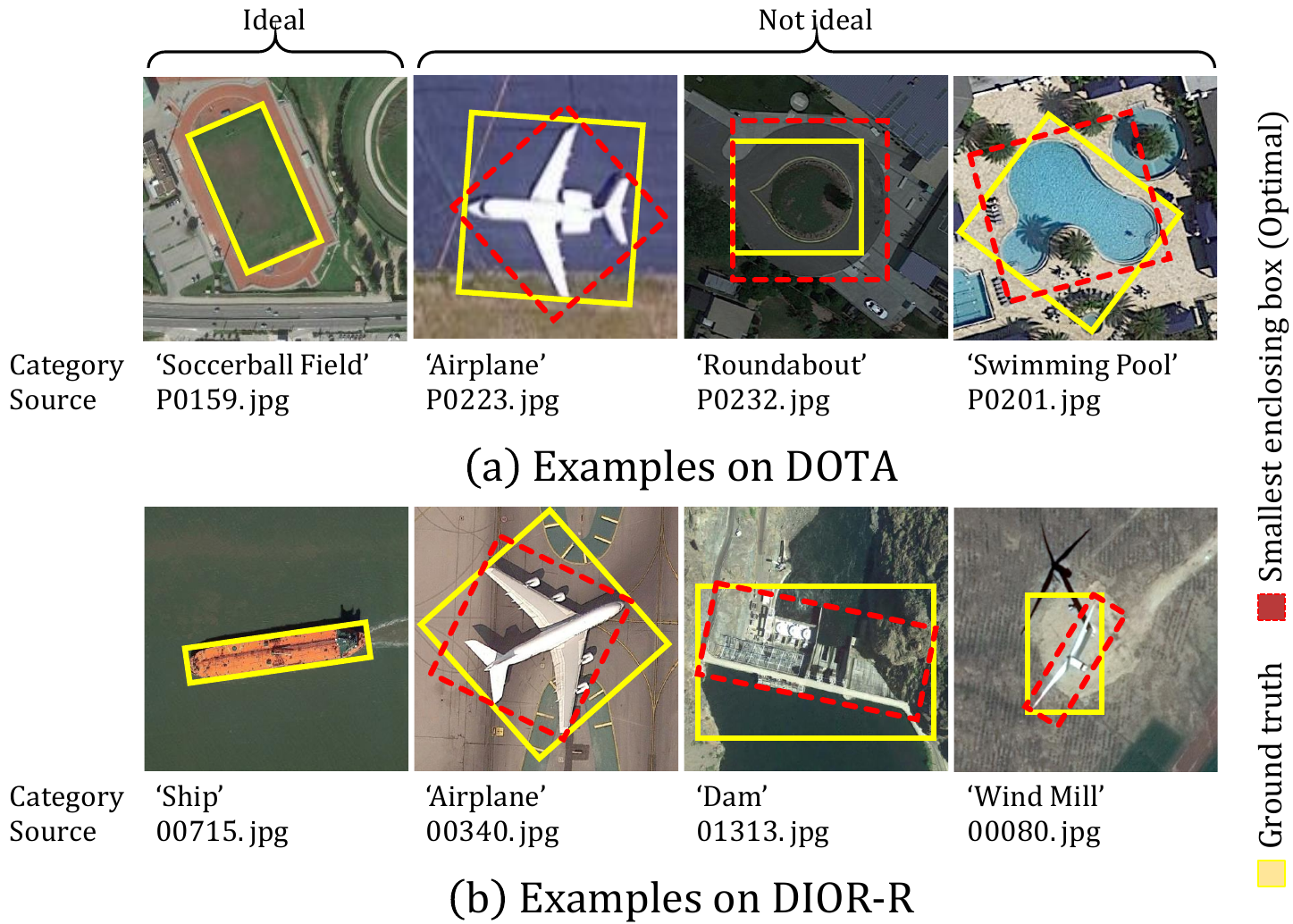}
    \caption{\small{Comparison between provided bounding box labels and minimum enclosing rectangles for objects on DOTA and DIOR-R.}}
    \label{fig2}
    \vspace{-0.3cm}
\end{figure}

\vspace{-0.2cm}
\section{Methodology}
In this section, we present a simple, but efficient data augmentation strategy namely \textbf{NBBOX} (\underline{N}oise Injection into \underline{B}ounding \underline{Box}) for remote sensing object detection.
    As seen in Fig. \ref{fig2}, there is often a discrepancy between the actual annotation and the minimum enclosing box (i.e. optimal localization label) on the existing remote sensing datasets, which might cause difficulties in model training due to inaccurate and low-consistent bounding boxes.
    To enhance the model's robustness against such implicit box-level noise, our augmentation scheme randomly adds noise to oriented bounding boxes with the three simplest geometric transformations: scaling, rotation, and translation during training, as depicted in Fig. \ref{fig1}.

\vspace{-0.3cm}
\subsection{Background}
Training models under label noise is challenging and requires careful consideration to learn resilient and generalized models.
    Most research attention has been given to the classification task, but recently a few have dedicated to object detection (e.g. \cite{chadwick2019training, li2020towards, xu2021training, liu2022robust} for natural scenes and \cite{wei2021object, bernhard2021correcting, bernhard2022robust} for remote sensing imagery).

For natural images, \cite{chadwick2019training} first analyzes the effects of various label noise types on object detection and introduces a per-object co-teaching framework to address the noise label issue. \cite{li2020towards} suggests a learning strategy of alternating between rectifying noise and training the model to handle label noise concerning category and bounding box. \cite{xu2021training} proposes to address label noise by adopting meta-learning via exploiting a few clean labels.
    Previous studies consider classification and localization simultaneously, whereas \cite{liu2022robust} employs multiple instance learning to deal solely with inaccurate bounding boxes. 
    We believe that it is reasonable and practical to pursue this research direction as imprecise bounding box labels are more prevalent than noisy category labels in practice.

For remote sensing images, a few work has been proposed to address noisy labels in object detection, inspired by research for natural scenes. 
    \cite{wei2021object} designs two kinds of loss functions to mitigate mislabels for both classification and localization.
    On the other hand, \cite{bernhard2021correcting}, \cite{bernhard2022robust} focus on correcting incomplete annotations such as missing bounding boxes due to misalignment of GPS sensors.
Unlike prior work, the main goal of our work is to resolve localization errors inherent in the existing bounding box labels. 
    In other words, we consider a generic object detection scenario, where missing bounding boxes or incorrect class labels of the ground truth are uncommon.
    Instead, we assume that the bounding box labels may not be optimal due to the annotators' mistakes or dilemma situations such as how to rotate a bounding box for a circular object on overhead imagery during the annotation process.
    In particular, we study a data augmentation strategy based on bounding box transformation for more robust localization.

\vspace{-0.3cm}
\subsection{Overview of the proposed approach}

The detail of \textbf{NBBOX} is presented in Algorithm \ref{algo1}.
    In our algorithm, there exist two types of input: $\mathbf{I}$ is the input image and $\mathbf{L}=\{(\mathbf{B}_{i}, \mathbf{c}_{i})\}^{N}_{i=1}$ is the corresponding labels of $N$ objects for object detection, where $\mathbf{B}_{i}=({x}_{\text{c}}, {y}_{\text{c}}, w, h, \theta)$ and $\mathbf{c}_{i}$ correspond to the bounding box label and the one-hot encoded category label for $i$-th object, respectively. $({x}_{\text{c}}, {y}_{\text{c}})$ is the center point, while $w$ and $h$ represent the width and height of the bounding box. $\theta$ denotes the rotated angle of the box.

Moreover, there are three kinds of hyper-parameters to tune. $\mathbf{s}=({s}_{\text{min}}, {s}_{\text{max}})$, $\mathbf{r}=({r}_{\text{min}}, {r}_{\text{max}})$, and $\mathbf{t}=({t}_{\text{min}}, {t}_{\text{max}})$ determine the ranges of translation, scaling, and rotation, respectively, where $\forall{k}, {s}_{k} \in \mathbb{R} \wedge {r}_{k} \in \mathbb{R} \wedge {t}_{k} \in \mathbb{Z}$. Here, $\mathbb{Z}$ is the set of all integers whereas $\mathbb{R}$ is for all real numbers.

    ${\text{rand}}_{\text{int}}(a, c) \sim {P}_{\text{int}}$ selects and returns any specific integer $c$ such that $a \le b \le c$. 
    ${\text{rand}}_{\text{float}}(a, c) \sim {P}_{\text{float}}$ generates any real number value within the closed interval $[a, c)$.
    We simply choose the uniform distributions for all random functions used in our algorithm. 
    In other words, ${P}(b; a, c)=\frac{1}{c-a+1}$ and ${P}(b; a, c)=\frac{1}{c-a}$ are used for the probability mass (discrete) function ${P}_{\text{int}}$ and the probability density (continuous) function ${P}_{\text{float}}$, respectively.
    One might question why $\mathbf{t}$ has a domain of integer values while $\mathbf{s}$ and $\mathbf{r}$ have domains of real numbers. This is because translation operations are conducted on a pixel basis, while scaling and rotation factors can be the real number.

Assume that a single image $\mathbf{I}$ is given with labels $\mathbf{L}$ to our algorithm.
    The ultimate goal is to perform bounding box transformations during training as a data augmentation strategy.
    To clarify, we randomly add noise to each bounding box $\mathbf{B}_{i}$ in $\mathbf{L}$ for each training epoch.
    In Algorithm \ref{algo1}, for each box $\mathbf{B}_{i}$, three kinds of transformations are applied in sequence (scaling: line 4-5, rotation: line 7-8, translation: line 10-11).
    Note that it is possible to activate only a subset of these sub-operations during the execution of \textbf{NBBOX}. For instance, you can simply choose not to use `rotation' for \textbf{NBBOX} if needed.

\begin{algorithm}[!t]         \small
        \label{algo1}
    \caption{Procedure of \textbf{NBBOX}.}
\SetKwInOut{Input}{Input}
\SetKwInOut{Initialization}{Parameters}
\SetKwInOut{Output}{Output}

\Input{Input image $\mathbf{I}$, 
Labels $\mathbf{L}=\{(\mathbf{B}_{i}, \mathbf{c}_{i})\}^{N}_{i=1}$}
\Initialization{Scaling range $\mathbf{s}=({s}_{\text{min}}, {s}_{\text{max}}) $, \\
Rotation range $\mathbf{r}=({r}_{\text{min}}, {r}_{\text{max}}) $, \\
Translation range $\mathbf{t}=({t}_{\text{min}}, {t}_{\text{max}}) $} 
\Output{Input image $\mathbf{I}$, Updated labels $\mathbf{L}^{*}$}    
\For{$i \leftarrow 1$ \textbf{to} $N$}{
    ${x}_{\text{c}}, {y}_{\text{c}}, w, h, \theta$ $\leftarrow$ $\mathbf{B}_{i}$\; 
    \textcolor{red}{\texttt{\small{// (1) BBOX Scaling}}} \textcolor{white}{\;}
        $s_w, s_h \leftarrow \text{rand}_{\text{float}}({s}_{\text{min}}, {s}_{\text{max}}), \text{rand}_{\text{float}}({s}_{\text{min}}, {s}_{\text{max}})$\; 
        $w^{*}, h^{*} \leftarrow w \times s_w, h \times s_h$\;
    \textcolor{red}{\texttt{\small{// (2) BBOX Rotation}}} \textcolor{white}{\;}
        $r_{\theta} \leftarrow \text{rand}_{\text{float}}({r}_{\text{min}}, {r}_{\text{max}})$\; 
        $\theta^{*} \leftarrow \theta + r_{\theta}$\; 
    \textcolor{red}{\texttt{\small{// (3) BBOX Translation}}} \textcolor{white}{\;}
        $t_x, t_y \leftarrow \text{rand}_{\text{int}}({t}_{\text{min}}, {t}_{\text{max}}), \text{rand}_{\text{int}}({t}_{\text{min}}, {t}_{\text{max}})$\; 
        ${x}^{*}_{\text{c}}, {y}^{*}_{\text{c}} \leftarrow {x}_{\text{c}} + t_x, {y}_{\text{c}} + t_y$\; 
    \textcolor{red}{\texttt{\small{// Update}}} \textcolor{white}{\;}
        ${\mathbf{B}}^{*}_{i} \leftarrow {x}^{*}_{\text{c}}, {y}^{*}_{\text{c}}, w^{*}, h^{*}, \theta^{*}$\; 
} 
${\mathbf{L}}^{*} \leftarrow \{(\mathbf{B}^{*}_{i}, \mathbf{c}_{i})\}^{N}_{i=1}$\; 
\vspace{-0.05cm}
\end{algorithm}


\vspace{-0.2cm}
\subsection{Advanced options for translation and scaling}
As demonstrated in Algorithm \ref{algo1}, the bounding box may be translated by different values along both the x-axis and y-axis. Besides, the aspect ratio may not be maintained for scaling operations.
    This phenomenon would not be recommended for training on certain data scenarios.
    For instance, it would not be desirable to modify the aspect ratio of bounding boxes in an extreme situation where most objects are expected to have a square size.
    Hence, we alleviate this by providing additional parameters ${\text{bool}}_{s}$ and ${\text{bool}}_{t}$ in our algorithm. 
    Here, ${\text{bool}}_{s}$ and ${\text{bool}}_{t}$ are boolean values to be set by users. 
    One can effortlessly modify \textbf{NBBOX} with ${\text{bool}}_{s}$ and ${\text{bool}}_{t}$ as follows.

    For scaling, line 4 should be replaced with:
    \begin{equation}\label{eq1}
        \begin{split}
            & \alpha, \beta \leftarrow {\text{rand}}_{\text{float}}({s}_{\text{min}}, {s}_{\text{max}}), {\text{rand}}_{\text{float}}({s}_{\text{min}}, {s}_{\text{max}});  \\
            & s_w \leftarrow \alpha \text{ if } {\text{bool}}_{s} \text{ is True, otherwise } \alpha;\\
            & s_h \leftarrow \alpha \text{ if } {\text{bool}}_{s} \text{ is True, otherwise } \beta,\\
        \end{split}
        \end{equation}
        where ${\text{bool}}_{s}$ represents whether the bounding box is proportionally resized to keep the width and height ratio unchanged.

    Likewise, for translation, line 10 should be replaced with:
    \begin{equation}\label{eq2}
        \begin{split}
            & \alpha, \beta \leftarrow {\text{rand}}_{\text{int}}({t}_{\text{min}}, {t}_{\text{max}}), {\text{rand}}_{\text{int}}({t}_{\text{min}}, {t}_{\text{max}});  \\
            & t_x \leftarrow \alpha \text{ if } {\text{bool}}_{t} \text{ is True, otherwise } \alpha;\\
            & t_y \leftarrow \alpha \text{ if } {\text{bool}}_{t} \text{ is True, otherwise } \beta,\\
        \end{split}
        \end{equation}
        where ${\text{bool}}_{t}$ denotes whether the center point is shifted equally along both the $x$ and $y$ directions.

\vspace{-0.2cm}
\subsection{Scale-aware noise injection into bounding box}
Tiny objects often occupy only a few pixels in an image, making their bounding boxes highly sensitive to even minor perturbations.
    For tiny objects, such transformations can lead to significant discrepancies between the transformed bounding box and the actual object, resulting in huge misalignment \cite{xu2022detecting}.
    To fill this gap, we propose enhancing our algorithm by replacing line 2 in Algorithm \ref{algo1} with:

\vspace{-0.2cm}
\begin{equation}\label{eq3}
        \begin{split}
            & {x}_{\text{c}}, {y}_{\text{c}}, w, h, \theta \leftarrow \mathbf{B}_{i};  \\
            & \text{if } w \le \gamma \text{ or } h \le \gamma: \text{continue},\\
        \end{split}
        \end{equation}
        where $\gamma \in \mathbb{Z}$ is a hyper-parameter to tune.
        For example, $\gamma=0$ means applying random transformations to all bounding boxes, as in our original algorithm.
        In contrast, when $\gamma$ is a positive integer, it indicates that bounding boxes with a width or height smaller than $\gamma$ will remain unaltered. 
         Hence, the proper value of $\gamma$ helps preserve the integrity of tiny objects, ensuring that their spatial properties are not compromised by transformations that may lead to misalignment or loss of information in remote sensing imagery.
         Note that our Python implementation with Numpy slightly deviates from our pseudo-code for efficient computation (See supp. material).

\section{Experiments}
In this section, we show our data augmentation strategy \textbf{NBBOX} improves the performance of remote sensing object detection. 
    We first briefly introduce the experimental settings and implementation details. 
    We then provide test results and analysis of experiments with the proposed method.
    \vspace{-0.2cm}
    
\subsection{Experimental setup}
\label{experimental-setup}
\textbf{Detection models}. Traditionally, deep learning-based object detection models are typically separated into one-stage and two-stage methods.
    Nowadays, the distinction between anchor-based and anchor-free methodologies tends to be more emphasized when classifying detector models.
    This work adopts rotated Faster R-CNN \cite{ren2016faster} and FCOS \cite{detector2022fcos}, representative models in anchor-based two-stage and anchor-free one-stage detection studies, respectively.
    For all models, either ImageNet-pretrained ResNet-50 \cite{he2016deep}, a representative CNN model, or Swin-T \cite{liu2021swin}, a representative Transformer model, is used for feature extraction.
    Additionally, FPN-1x \cite{lin2017feature} is incorporated as a neck to enhance multi-scale invariance.

\textbf{Training details}. We implement our hypothesis using PyTorch and MMRotate \cite{zhou2022mmrotate}. 
    All the models are trained for 25 epochs.
    For a fair comparison with other approaches, all the experiments are resumed with weights pre-trained for five epochs without any augmentation per each \{detector, data\} combination.
    SGD is used for optimization with momentum of weight 0.9, weight decay 1e-4, learning rate 1e-3, and batch size 8.
    Learning rate warm-up is used over the first 500 iterations, where the learning rate linearly increases from zero to 33\% of 5e-3.
    L2-norm gradient clipping is utilized with a maximum norm of 35.
    We apply the pre-processing operations `RResize', `RRandomFlip', `Normalize', and `Pad' for training, while `MultiScaleFlipAug' is also used for testing.

\begin{figure*}[ht!]
    \centering
    \includegraphics[width=16.5cm]{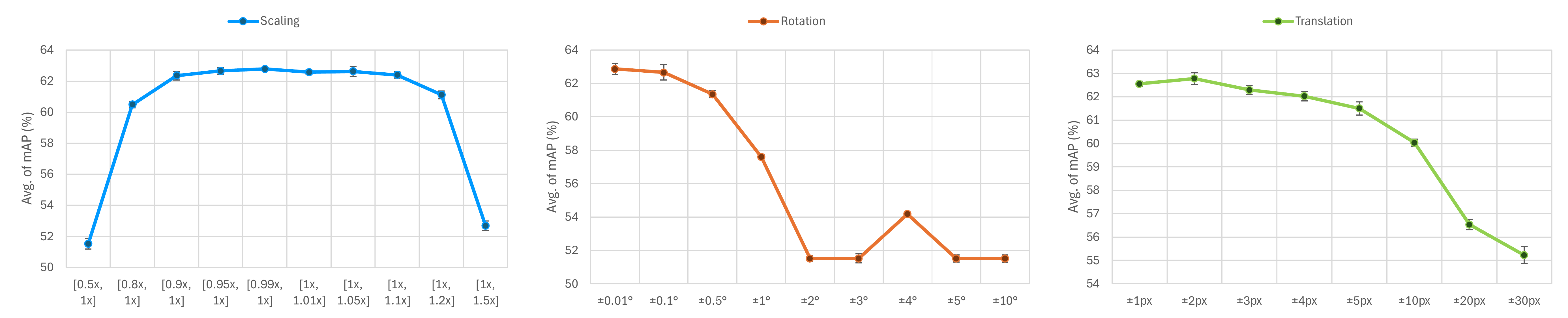} \vspace{-0.1cm}
    \caption{\small{Impact of scaling, rotation, and translation of bounding boxes on model performance: as we observe similar outcomes across different architectures and datasets, for brevity, we include only the result from Faster R-CNN with ResNet-50 trained on DIOR-R.}}
    \label{fig3}
   \vspace{-0.3cm}
\end{figure*}

\textbf{Evaluation metric}. To objectively verify performance, all experiments are repeated five times, and the evaluation results are reported as `mean ± standard deviation'.
    For each experiment, we use mAP (mean Average Precision) at IoU($\hat{\textbf{y}}, \textbf{y}$)=0.5 (i.e. mAP@50) to measure the accuracy of the detected bounding boxes ($\hat{\textbf{y}}$) against the ground truth boxes ($\textbf{y}$) following \cite{lin2014microsoft}. 
    This metric evaluates both the precision and recall of the model across different classes as:
    \begin{equation}\label{eq4}
        \textstyle \text{mAP} = \frac{1}{C} \sum_{i=1}^{C} {\text{AP}}_{i},
        \end{equation}
        where $C$ is the number of categories and ${\text{AP}}_{i}$ is the average precision for category $i$. See \cite{wang2022parallel} for detailed calculation.

\vspace{-0.3cm}
\subsection{Datasets}
    This work conducts experiments on two widely recognized datasets for remote sensing object detection: DOTA (v1) \cite{xia2018dota} and DIOR-R \cite{cheng2022anchor}. 
        Both have diverse real-world categories, high resolution, and detailed annotations.
        DOTA includes 15 object categories such as `car', `ship', and `harbor'.
        It contains 2,806 aerial images of varying resolutions with 188,282 oriented bounding box labels.
        Compared to DOTA, DIOR-R has 20 object categories including 192,512 instances over 23,463 images for rotated object detection.
        We conduct experiments and report results based on the standard convention like \cite{li2023instance} in the remote sensing community for both datasets.

\vspace{-0.3cm}
\subsection{Impact of scaling, rotation, and translation in \textbf{NBBOX}}
\label{isrt}

This subsection studies the model performance when exposed to varying levels of scaling, rotation, and translation of bounding boxes.
    One fact is that too much deformation will adversely affect model training.
    In other words, to mitigate the bad effects of \textbf{NBBOX}, we assume that (i) ${s}_{\text{min}}$ and ${s}_{\text{max}}$ should be close to 1; 
                                                            (ii) ${r}_{\text{min}}$ and ${r}_{\text{max}}$ should be close to 0; 
                                                            (iii) ${t}_{\text{min}}$ should be close to -1 and ${t}_{\text{max}}$ should be close to 1.
    This reduces the difficulty of parameter tuning. 
    To validate this assumption, we conduct experiments on DIOR-R with Faster R-CNN and FCOS, both employing ResNet-50 as the backbone (See Fig. \ref{fig3} and supp. material for details).
    Here we consider combinations of the following parameters: ${s}_{\text{min}} \in$ \{0.5, 0.8, 0.9, 0.95, 0.99\},  ${s}_{\text{max}} \in$ \{1, 1.01, 1.05, 1.1, 1.2, 1.5\},  $\{{r}_{\text{min}}, {r}_{\text{max}}\} \in$ \{±0.01, ±0.1, ±0.5, ±1, ±2, ±3, ±4, ±5, ±10\}, and $\{{t}_{\text{min}}, {t}_{\text{max}}\} \in$ \{±1, ±2, ±3, ±4, ±5, ±10, ±20, ±30\}. 
As a result, we find parameters that support our hypothesis as:
\begin{itemize} \small
    \item ${s}_{\text{min}} \in$ \{0.95, 1\}, ${s}_{\text{max}} \in$ \{1, 1.01\};
    \item $\{{r}_{\text{min}}, {r}_{\text{max}}\}$ = \{±0.01\} \& ${t}_{\text{min}} \in$ \{-1, -2\}, ${t}_{\text{max}} \in$ \{1, 2\}.
\end{itemize}
Unless otherwise specified, we use ${s}_{\text{min}}$ = 0.99, ${s}_{\text{max}}$ = 1.01, $\{{r}_{\text{min}}, {r}_{\text{max}}\}$ = \{±0.01\}, and $\{{t}_{\text{min}}, {t}_{\text{max}}\}$ = \{±1\} for remaining.

\textbf{Isotropic scaling and translation}. When scaling with preserving aspect ratio (i.e. ${\text{bool}}_{s}$=True), all models show a slight increase in mAP compared to when it is not (i.e. ${\text{bool}}_{s}$=False) as shown in Table \ref{tab2}.
    Besides, in the case of translation noise, the models perform slightly better when isotropic translation is applied (i.e. ${\text{bool}}_{t}$=True) compared to when it is not (i.e. ${\text{bool}}_{t}$=False) as indicated in Table \ref{tab2}. 
    Thus, experimental results on DIOR-R in Table \ref{tab2} suggest that isotropic scaling and translation may be important.
    In the remaining experiments, we use ${\text{bool}}_{s}$=True and ${\text{bool}}_{t}$=True.

\begin{table}[h!]
\vspace{-0.3cm}
\centering
\caption{\small{\label{tab:table-name} Average mAP under isotropic/non-isotropic scaling and \\translation noises for each parameters \\(${s}_{\text{min}}$, ${s}_{\text{max}}$, ${t}_{\text{min}}$, and ${t}_{\text{max}}$) on DIOR-R in \textbf{NBBOX}.}} \vspace{-0.2cm}
\resizebox{5cm}{!}{%
\begin{tabular}{@{}cc|ccc@{}}
\toprule
\multicolumn{2}{c|}{(With ResNet-50)} & Faster R-CNN & FCOS \\ \midrule
\multirow{2}{*}{Scaling} & ${\text{bool}}_{s}$=True & \textbf{60.12} (\textcolor{red}{+0.05}) & \textbf{56.97} (\textcolor{red}{+0.05})  \\
 & ${\text{bool}}_{s}$=False & 60.07 \textcolor{white}{(+0.00)} & 56.92 \textcolor{white}{(+0.00)} \\ \midrule
\multirow{2}{*}{Translation} & ${\text{bool}}_{t}$=True & \textbf{60.37} (\textcolor{red}{+1.08}) & \textbf{58.38} (\textcolor{red}{+1.46})  \\
 & ${\text{bool}}_{t}$=False & 59.29 \textcolor{white}{(+0.00)} & 56.92 \textcolor{white}{(+0.00)} \\ \bottomrule
\end{tabular}%
} \vspace{-0.2cm}
\label{tab2}
\end{table}

\begin{table}[ht!]
\centering
\caption{\small{\label{tab:table-name} Impact of combining scaling, rotation, and translation of bounding boxes on DIOR-R in \textbf{NBBOX}. \\\tiny(\textbf{Bold}: best, \underline{underlined}: second-best, \uwave{wavy-underlined}: third-best in this paper)}} 
\vspace{-0.2cm}
\resizebox{5.7cm}{!}{%
\begin{tabular}{@{}c|ccc@{}}
\toprule
(With ResNet-50) & Faster R-CNN & FCOS \\ \midrule
Scaling (Best) & 62.79 ± 0.05 & 60.00 ± 0.99  \\
Rotation (Best) & 62.87 ± 0.29 & 59.68 ± 1.03 \\
Translation (Best) & 62.78 ± 0.26 & 59.93 ± 0.91 \\
Scaling + Rotation & \underline{62.92} ± 0.24 & \underline{60.13} ± 0.97 \\
Scaling + Translation & \uwave{62.89} ± 0.30 & \uwave{60.02} ± 1.07  \\
Translation + Rotation & 62.80 ± 0.22 & 59.89 ± 1.06  \\
Scaling + Rotation + Translation & \textbf{63.18} ± 0.14 & \textbf{60.25} ± 1.00  \\ \bottomrule
\end{tabular}%
} \vspace{-0.2cm}
\label{tab3}
\end{table}

\begin{table}[ht!]
\centering
\caption{\small{\label{tab:table-name} Impact of injecting scale-aware noise on DIOR-R in \textbf{NBBOX}.}} 
\vspace{-0.2cm}
\resizebox{8cm}{!}{%
\begin{tabular}{@{}cccccc@{}}
\toprule
(With ResNet-50)    & $\gamma$ = 0 & $\gamma$ = 8 & $\gamma$ = 16 & $\gamma$ = 32 & $\gamma$ = 64 \\ \midrule
Faster R-CNN         & \uwave{63.18} ± 0.14 & \underline{63.20} ± 0.20 & \textbf{63.21} ± 0.18 & 63.17 ± 0.17 & 62.93 ± 0.27 \\
FCOS                & \underline{60.25} ± 1.00 & \uwave{60.24} ± 1.18 & \textbf{60.27} ± 1.06 & 59.99 ± 1.30 & 59.74 ± 1.24 \\ \bottomrule
\end{tabular}%
} \vspace{-0.2cm}
\label{tab5}
\end{table}

\textbf{Combining scaling, rotation, and translation}. There is a reasonable expectation that integrating scaling, rotation, and translation could synergistically improve the model's robustness and generalization.
    Hence, we further conduct experiments to validate the effectiveness of such combinations.
    Precisely, we study different combinations of each transformation (with parameters that show the best): pairwise combinations and the complete combination.
    Table \ref{tab3} indicates the effectiveness of the combined transformations in enhancing the model's performance. 
    Notably, the improvements are particularly evident in the scenario when we simultaneously use scaling, rotation, and translation during training.

\vspace{-0.2cm}
\subsection{Impact of scale-aware noise injection in \textit{\textbf{NBBOX}}}
Following \ref{isrt}, this subsection particularly focuses on a parameter study of $\gamma$, where $\gamma$ is varied over the set \{0, 8, 16, 32, 64\} under the combination of scaling, rotation, and translation.
    For this, we perform experiments on DIOR-R using Faster R-CNN and FCOS, both configured with ResNet-50 as the backbone as in \ref{isrt}. 
    As demonstrated in Table \ref{tab5}, both detectors achieve their highest accuracies at $\gamma=16$, suggesting that bounding box perturbation for tiny objects (less than 16 $\times$ 16 pixels) might not be recommended. 
    Besides, as $\gamma$ increases beyond 16, the performance degrades, indicating that moderate noise injection can enhance the model's robustness.
    Hence, we use $\gamma$ = 16 for the remaining experiments.

\subsection{Comparison with state-of-the-art methods}
In this subsection, we compare our method with easy-to-get data augmentation methods based on image manipulation (image/patch erasing and mix) as illustrated in Table \ref{tab4}. 
For a DOTA dataset, we use the same hyper-parameters for DIOR-R.
Table \ref{tab4} shows that our \textbf{NBBOX} achieves a competitive performance among the state-of-the-art augmentation methods.
    Its performance can be further enhanced with other methods, such as RandRotate. 
    Notably, obtaining satisfactory results on the DOTA dataset using the same parameters derived from DIOR-R demonstrates that the proposed method requires minimal parameter tuning.
    In addition, our method is faster than others: training Faster R-CNN with ResNet-50 on DIOR-R takes an average of 9.63 minutes per epoch for our method, compared to 21.24 minutes for MosaicMix.
    \textcolor{black}{Due to the page limit, more results and analysis are in the supp. material.}

\begin{table}[ht!]
\vspace{-0.3cm}
\centering
\caption{\small{\label{tab:table-name} Comparison of state-of-the-art methods.}} \vspace{-0.2cm}
\resizebox{7.8cm}{!}{%
\begin{tabular}{@{}c|l|cccc@{}}
\toprule
\multirow{2.5}{*}{Backbone} & \multicolumn{1}{c|}{Data} & \multicolumn{2}{c|}{DIOR-R} & \multicolumn{2}{c}{DOTA} \\ \cmidrule(l){2-6} 
 & \multicolumn{1}{c|}{Detectors} & Faster R-CNN & \multicolumn{1}{c|}{FCOS} & Faster R-CNN & FCOS \\ \midrule
\multirow{15}{*}{\rotatebox{90}{ResNet-50}} & Baseline & 62.66 ± 0.17 & \multicolumn{1}{c|}{59.75 ± 1.09} & 68.92 ± 0.44 & 67.00 ± 0.26 \\
 & + RandRotate \cite{lalitha2022review} & \underline{63.10} ± 0.21 & \multicolumn{1}{c|}{\underline{60.11} ± 1.12} & \underline{70.02} ± 0.32 & \underline{67.56} ± 0.49 \\
 & + RandShift \cite{lalitha2022review} & 62.87 ± 0.37 & \multicolumn{1}{c|}{59.67 ± 1.05} & 69.01 ± 0.40 & 67.22 ± 0.59 \\
 & + CutOut \cite{devries2017improved} & \uwave{62.91} ± 0.26 & \multicolumn{1}{c|}{59.84 ± 0.99} & 69.13 ± 0.45 & 67.42 ± 0.37 \\
 & + AutoAug \cite{cubuk2019autoaugment} & 62.33 ± 0.31 & \multicolumn{1}{c|}{59.24 ± 1.07} & 68.87 ± 0.32 & \uwave{67.45} ± 0.49 \\
 & + RandAug \cite{cubuk2020randaugment} & 62.55 ± 0.12 & \multicolumn{1}{c|}{\uwave{59.94} ± 1.29} & 66.11 ± 0.24 & 65.08 ± 0.70 \\
 & + AugMix \cite{hendrycks2020augmix} & 62.83 ± 0.11 & \multicolumn{1}{c|}{59.93 ± 0.75} & 69.26 ± 0.10 & 66.71 ± 0.52 \\
 & + MosaicMix \cite{takahashi2019data, wei2020amrnet} & 61.15 ± 0.19 & \multicolumn{1}{c|}{58.03 ± 0.76} & 62.73 ± 0.58 & 61.18 ± 0.57 \\
 & + UniformAug \cite{lingchen2020uniformaugment} & 59.18 ± 0.18 & \multicolumn{1}{c|}{56.35 ± 1.53} & 65.61 ± 0.37 & 64.35 ± 0.84 \\
  & + TrivialAug \cite{muller2021trivialaugment} & 60.22 ± 0.17 & \multicolumn{1}{c|}{57.37 ± 1.48} & 66.54 ± 0.19 & 65.34 ± 0.56 \\
 & + YOCO \cite{han2022you} & 57.52 ± 0.16 & \multicolumn{1}{c|}{54.37 ± 1.70} & 65.16 ± 0.36 & 63.12 ± 0.48 \\
 & + PixMix \cite{hendrycks2022pixmix} & 62.07 ± 0.23 & \multicolumn{1}{c|}{58.91 ± 1.13} & 68.72 ± 0.33 & 66.82 ± 0.28 \\
 & + InterAug \cite{devi2023improving} & 60.94 ± 0.27 & \multicolumn{1}{c|}{58.45 ± 1.20} & 66.93 ± 0.24 & 66.52 ± 0.71 \\
 & + \textbf{Ours} & \textbf{63.21} ± 0.18 & \multicolumn{1}{c|}{\textbf{60.27} ± 1.06} & \uwave{69.78} ± 0.54 & \textbf{67.89} ± 0.46 \\
 & + \textbf{Ours + RandRotate} & - & \multicolumn{1}{c|}{-} & \textbf{70.37} ± 0.13 & - \\ \midrule
\multirow{17}{*}{\rotatebox{90}{Swin-T}}& Baseline & 61.53 ± 0.49 & \multicolumn{1}{c|}{63.60 ± 0.99} & 71.03 ± 0.47 & 68.40 ± 0.68 \\
 & + RandRotate \cite{lalitha2022review} & 62.54 ± 0.43 & \multicolumn{1}{c|}{\underline{64.08} ± 0.78} & \underline{71.62} ± 0.34 & 68.52 ± 0.95 \\
 & + RandShift \cite{lalitha2022review} & \uwave{62.77} ± 0.30 & \multicolumn{1}{c|}{63.22 ± 0.73} & 71.21 ± 0.50 & 68.51 ± 0.64 \\
 & + Cutout \cite{devries2017improved} & \underline{63.17} ± 0.36 & \multicolumn{1}{c|}{\uwave{63.80} ± 0.88} & 71.29 ± 0.28 & 68.41 ± 0.82 \\
 & + AutoAug \cite{cubuk2019autoaugment} & 62.80 ± 0.43 & \multicolumn{1}{c|}{63.19 ± 0.94} & 70.78 ± 0.64 & 69.04 ± 0.72 \\
 & + RandAug \cite{cubuk2020randaugment} & 60.54 ± 0.60 & \multicolumn{1}{c|}{62.07 ± 0.95} & 68.63 ± 0.41 & 68.91 ± 0.68 \\
 & + AugMix \cite{hendrycks2020augmix} & 62.55 ± 0.27 & \multicolumn{1}{c|}{63.07 ± 1.01} & 70.48 ± 0.61 & \underline{69.06} ± 0.57 \\
 & + MosaicMix \cite{takahashi2019data, wei2020amrnet} & 58.26 ± 0.31 & \multicolumn{1}{c|}{60.57 ± 0.76} & 66.48 ± 0.39 & 67.24 ± 0.72 \\
   & + UniformAug \cite{lingchen2020uniformaugment} & 58.48 ± 0.21& \multicolumn{1}{c|}{60.55 ± 0.91} & 67.38 ± 0.45 & \uwave{69.06} ± 0.71\\
 & + TrivialAug \cite{muller2021trivialaugment} & 59.72 ± 0.27& \multicolumn{1}{c|}{61.44 ± 1.12} & 67.63 ± 0.42 & 68.53 ± 0.69\\
 & + YOCO \cite{han2022you} & 57.05 ± 0.32& \multicolumn{1}{c|}{58.79 ± 1.09} & 66.26 ± 0.39& 69.04 ± 0.59\\
 & + PixMix \cite{hendrycks2022pixmix} & 58.15 ± 0.31& \multicolumn{1}{c|}{58.22 ± 0.81} & 66.90 ± 0.60& 68.15 ± 0.32\\
 & + InterAug \cite{devi2023improving} & 60.72 ± 0.20 & \multicolumn{1}{c|}{62.24 ± 0.99} & 69.35 ± 0.38 & 67.09 ± 0.84 \\
 & + \textbf{Ours} & 62.22 ± 0.23 & \multicolumn{1}{c|}{63.64 ± 0.89} & \uwave{71.30} ± 0.35 & 68.94 ± 0.59\\
 & + \textbf{Ours + Cutout} & \textbf{63.18} ± 0.30 & \multicolumn{1}{c|}{-} & - & - \\
 & + \textbf{Ours + RandRotate} & - & \multicolumn{1}{c|}{\textbf{64.11} ± 0.98} & \textbf{71.69} ± 0.41 & - \\
 & + \textbf{Ours + AugMix} & - & \multicolumn{1}{c|}{-} & - & \textbf{69.12} ± 0.50 \\
 \bottomrule
\end{tabular}%
} \vspace{-0.3cm}
\label{tab4}
\end{table}

\vspace{-0.2cm}
\section{Conclusion}
In this letter, we have proposed a novel data augmentation method named \textbf{NBBOX} for remote sensing object detection.
    This work thoroughly investigates the effect of bounding box transformation with scaling, rotation, and translation for remote sensing object detection.
    Besides, \textbf{NBBOX} is easy to integrate with modern deep-learning frameworks and more time-efficient than other augmentation methods.
    In the future, we will extend our method with various AI paradigms.
    \vspace{-0.15cm}

\ifCLASSOPTIONcaptionsoff
  \newpage
\fi



%


\newpage

\lstset{
  language=Python,
  basicstyle=\ttfamily\tiny,      
  breaklines=true,                 
  breakatwhitespace=true,          
  keywordstyle=\color{blue},      
  commentstyle=\color{gray},      
  stringstyle=\color{teal},       
  numbers=left,                   
  numberstyle=\tiny\color{gray},  
  stepnumber=1,                   
  numbersep=8pt,                  
  backgroundcolor=\color{white},  
  showspaces=false,               
  showstringspaces=false,         
  showtabs=false,                 
  frame=lines,                    
  tabsize=2,                      
  captionpos=b,                   
  breaklines=true,                
  breakatwhitespace=false,        
  escapeinside={\%*}{*)},         
  lineskip=1pt
}

\onecolumn
\appendix

\subsection{Source code availability}
In this work, we release our code implementation of \textbf{NBBOX} (\underline{N}oise Injection into \underline{B}ounding \underline{Box}) on GitHub at \url{https://github.com/unique-chan/NBBOX} to help researchers build upon our efforts and replicate our experimental results.
    Please refer to the `readme.md' file in the GitHub repository for better usage.
    We also present our code implementation for the proposed algorithm in Python3 / Numpy / MMDetection (or MMRotate) as follows:

\vspace{0.2cm}

\noindent\makebox[\textwidth] {
\centering
\begin{minipage}{18cm}
\label{algo1}
\begin{lstlisting}[label=algo1,caption=\small\textsc{Python implementation of the proposed method named \textbf{NBBOX}}.]
class NoisyBBOX:
    def __init__(self, scale_range=(0.99, 1.01), isotropically_rescaled=True,
                 angle_range=(-0.01, 0.01), translate_range=(-1, 1), isotropically_translated=True,
                 threshold=16):
        self.scale_range = scale_range
        self.isotropically_rescaled = isotropically_rescaled
        self.angle_range = angle_range
        self.translate_range = translate_range
        self.isotropically_translated = isotropically_translated
        self.threshold = threshold

    def __call__(self, results):
        bboxes = results['ann_info']['bboxes']
        # (1) BBOX Scaling
        if self.isotropically_rescaled:
            scales = np.random.uniform(self.scale_range[0], self.scale_range[1], size=(bboxes.shape[0]))
            scales[(bboxes[:, 2] < self.threshold) | (bboxes[:, 3] < self.threshold)] = 1
            bboxes[:, 2] *= scales; bboxes[:, 3] *= scales
        else:
            scales = np.random.uniform(self.scale_range[0], self.scale_range[1], size=(bboxes.shape[0],2))
            scales[(bboxes[:, 2] < self.threshold) | (bboxes[:, 3] < self.threshold)] = 1
            bboxes[:, 2] *= scales[:, 0]; bboxes[:, 3] *= scales[:, 1]
        # (2) BBOX Rotation
        angles = np.random.uniform(self.angle_range[0], self.angle_range[1], size=(bboxes.shape[0]))
        angles[(bboxes[:, 2] < self.threshold) | (bboxes[:, 3] < self.threshold)] = 0
        bboxes[:, 4] += angles
        # (3) BBOX Translation
        if self.translate_range[0] < self.translate_range[1]:
            if self.isotropically_translated:
                translations = np.random.randint(self.translate_range[0], self.translate_range[1],
                                                 size=(bboxes.shape[0]))
                translations[(bboxes[:, 2] < self.threshold) | (bboxes[:, 3] < self.threshold)] = 0
                bboxes[:, 0] += translations; bboxes[:, 1] += translations
            else:
                translations = np.random.randint(self.translate_range[0], self.translate_range[1],
                                                 size=(bboxes.shape[0],2))
                translations[(bboxes[:, 2] < self.threshold) | (bboxes[:, 3] < self.threshold), :] = 0
                bboxes[:, 0] += translations[:, 0]; bboxes[:, 1] += translations[:, 1]
        # Update
        results['ann_info']['bboxes'] = bboxes
        if 'img_info' in results and 'ann' in results['img_info']:
            results['img_info']['ann']['bboxes'] = bboxes
        results['gt_bboxes'] = results['ann_info']['bboxes'].copy()
        return results
\end{lstlisting}
\end{minipage}
}
\vspace{0.2cm}

In the above code, \texttt{self.isotropically\_rescaled} and \texttt{self.isotropically\_translated} correspond to ${\text{bool}}_{s}$ and ${\text{bool}}_{t}$, respectively (See II-C. in the main manuscript for details).
    In addition, masking codes (in lines \{17, 21, 25, 32, 37\}) are implemented for scale-aware noise injection (See II-D. in the main manuscript for details).
    For instance, if the width or height of a bounding box is below \texttt{self.threshold} ($\gamma$), that box is not translated, as, in line 32 (or 37), the translation value for the corresponding box is replaced with 0.
    Refer to \url{https://mmdetection.readthedocs.io/en/v2.28.2/tutorials/data_pipeline.html} for understanding user-customized data augmentations in the detection framework based on MMDetection (or MMRotate).

\subsection{Experimental environments}
All experiments in this work have been conducted on a single machine equipped with AMD Ryzen 7 3700x (CPU), NVIDIA RTX 3090 (GPU) and Ubuntu 20.04 (OS).

\subsection{Parameter study of scaling, rotation, and translation in \textbf{NBBOX}.}
This subsection studies the model performance when exposed to varying levels of scaling, rotation, and translation of bounding boxes in \textbf{NBBOX}.
    For this, we consider combinations of the following parameters: ${s}_{\text{min}} \in$ \{0.5, 0.8, 0.9, 0.95, 0.99\},  ${s}_{\text{max}} \in$ \{1, 1.01, 1.05, 1.1, 1.2, 1.5\}, ${\text{bool}}_{s} \in$ \{\text{True}, \text{False}\},  $\{{r}_{\text{min}}, {r}_{\text{max}}\} \in$ \{±0.01, ±0.1, ±0.5, ±1, ±2, ±3, ±4, ±5, ±10\}, and  $\{{t}_{\text{min}}, {t}_{\text{max}}\} \in$ \{±1, ±2, ±3, ±4, ±5, ±10, ±20, ±30\}, ${\text{bool}}_{t} \in$ \{\text{True}, \text{False}\}. 
    Due to time limits, we conduct experiments only on DIOR-R with Faster R-CNN and FCOS, both employing ResNet-50 as the backbone.
    As demonstrated in Table \ref{tab1}, following hyper-parameters are recommended: ${s}_{\text{min}} \in$ \{0.95, 1\}, ${s}_{\text{max}} \in$ \{1, 1.01\}; $\{{r}_{\text{min}}, {r}_{\text{max}}\}$ = \{±0.01\}; ${t}_{\text{min}} \in$ \{-1, -2\}, ${t}_{\text{max}} \in$ \{1, 2\}.
    These results support our assumption in Section III-D of the main manuscript.
    DIOR-R is one of the challenging remote sensing datasets that, while having a relatively small number of images, contains samples across a diverse range of categories.
    Thus, we apply the best results of the parameter study conducted on DIOR-R directly to a DOTA dataset as well.
    Unless otherwise specified, we use ${s}_{\text{min}}$ = 0.99, ${s}_{\text{max}}$ = 1.01, $\{{r}_{\text{min}}, {r}_{\text{max}}\}$ = \{±0.01\}, and $\{{t}_{\text{min}}, {t}_{\text{max}}\}$ = \{±1\} for remaining.

\begin{table}[ht!]
\centering
\caption{\small{\label{tab:table-name} Parameter study of scaling, rotation, and translation of bounding boxes across various detection architectures on DIOR-R with \textbf{NBBOX} (Feature backbone: ResNet-50). \\\small(\textbf{Bold}: best, \underline{underlined}: second-best, \uwave{wavy-underlined}: third-best in this supp. material)}}
\vspace{-0.2cm}
\resizebox{7.4cm}{!}{%

\begin{tabular}{@{}cccccc@{}}
\toprule
\multicolumn{4}{c}{Configuration} & \multicolumn{2}{c}{Detectors} \\ \midrule
\multicolumn{1}{c|}{Transform} & \begin{tabular}[c]{@{}c@{}}${s}_{\text{min}}$\\ (${r}_{\text{min}}$)\\ <${t}_{\text{min}}$>\end{tabular} & \begin{tabular}[c]{@{}c@{}}${s}_{\text{max}}$\\ (${r}_{\text{max}}$)\\ <${t}_{\text{max}}$>\end{tabular} & \multicolumn{1}{c|}{\begin{tabular}[c]{@{}c@{}}${\text{bool}}_{s}$\\ <${\text{bool}}_{t}$>\end{tabular}} & Faster R-CNN & FCOS \\ \midrule
\multicolumn{1}{c|}{\multirow{20}{*}{Scaling}} & \multirow{2}{*}{0.5x} & \multirow{2}{*}{1x} & \multicolumn{1}{c|}{True} & 51.52 ± 0.34 & 42.17 ± 1.04 \\
\multicolumn{1}{c|}{} &  &  & \multicolumn{1}{c|}{False} & 51.52 ± 0.34 & 43.41 ± 1.17 \\
\multicolumn{1}{c|}{} & \multirow{2}{*}{0.8x} & \multirow{2}{*}{1x} & \multicolumn{1}{c|}{True} & 60.49 ± 0.21 & 58.50 ± 0.88 \\
\multicolumn{1}{c|}{} &  &  & \multicolumn{1}{c|}{False} & 60.52 ± 0.45 & 58.52 ± 1.03 \\
\multicolumn{1}{c|}{} & \multirow{2}{*}{0.9x} & \multirow{2}{*}{1x} & \multicolumn{1}{c|}{True} & 62.36 ± 0.28 & 59.60 ± 1.01 \\
\multicolumn{1}{c|}{} &  &  & \multicolumn{1}{c|}{False} & 62.39 ± 0.22 & 59.39 ± 0.76 \\
\multicolumn{1}{c|}{} & \multirow{2}{*}{0.95x} & \multirow{2}{*}{1x} & \multicolumn{1}{c|}{True} & \underline{62.66} ± 0.20 & \uwave{59.78} ± 0.78 \\
\multicolumn{1}{c|}{} &  &  & \multicolumn{1}{c|}{False} & 62.45 ± 0.11 & 59.58 ± 0.92 \\
\multicolumn{1}{c|}{} & \multirow{2}{*}{0.99x} & \multirow{2}{*}{1x} & \multicolumn{1}{c|}{True} & \textbf{62.79} ± 0.05 & \textbf{60.00} ± 0.99 \\
\multicolumn{1}{c|}{} &  &  & \multicolumn{1}{c|}{False} & \uwave{62.65} ± 0.18 & 59.74 ± 1.13 \\
\multicolumn{1}{c|}{} & \multirow{2}{*}{1x} & \multirow{2}{*}{1.01x} & \multicolumn{1}{c|}{True} & 62.57 ± 0.14 & \underline{59.91} ± 0.86 \\
\multicolumn{1}{c|}{} &  &  & \multicolumn{1}{c|}{False} & 62.71 ± 0.27 & 59.68 ± 1.09 \\
\multicolumn{1}{c|}{} & \multirow{2}{*}{1x} & \multirow{2}{*}{1.05x} & \multicolumn{1}{c|}{True} & 62.63 ± 0.32 & 59.68 ± 0.99 \\
\multicolumn{1}{c|}{} &  &  & \multicolumn{1}{c|}{False} & 62.28 ± 0.09 & 59.55 ± 1.12 \\
\multicolumn{1}{c|}{} & \multirow{2}{*}{1x} & \multirow{2}{*}{1.1x} & \multicolumn{1}{c|}{True} & 62.41 ± 0.21 & 59.56 ± 1.15 \\
\multicolumn{1}{c|}{} &  &  & \multicolumn{1}{c|}{False} & 62.12 ± 0.21 & 59.48 ± 0.97 \\
\multicolumn{1}{c|}{} & \multirow{2}{*}{1x} & \multirow{2}{*}{1.2x} & \multicolumn{1}{c|}{True} & 61.12 ± 0.24 & 58.77 ± 0.99 \\
\multicolumn{1}{c|}{} &  &  & \multicolumn{1}{c|}{False} & 61.03 ± 0.22 & 58.58 ± 0.95 \\
\multicolumn{1}{c|}{} & \multirow{2}{*}{1x} & \multirow{2}{*}{1.5x} & \multicolumn{1}{c|}{True} & 52.68 ± 0.32 & 51.72 ± 1.14 \\
\multicolumn{1}{c|}{} &  &  & \multicolumn{1}{c|}{False} & 53.12 ± 0.21 & 51.28 ± 0.90 \\ \midrule
\multicolumn{1}{c|}{\multirow{9}{*}{(Rotation)}} & -0.01$^{\circ}$ & 0.01$^{\circ}$ & \multicolumn{1}{c|}{-} & \textbf{62.87} ± 0.29 & \textbf{59.68} ± 1.03 \\
\multicolumn{1}{c|}{} & -0.1$^{\circ}$ & 0.1$^{\circ}$ & \multicolumn{1}{c|}{-} & \underline{62.66} ± 0.15 & \underline{59.66} ± 1.03 \\
\multicolumn{1}{c|}{} & -0.5$^{\circ}$ & 0.5$^{\circ}$ & \multicolumn{1}{c|}{-} & \uwave{61.35} ± 0.21 & \uwave{58.46} ± 0.91 \\
\multicolumn{1}{c|}{} & -1$^{\circ}$ & 1$^{\circ}$ & \multicolumn{1}{c|}{-} & 57.60 ± 0.26 & 54.16 ± 0.73 \\
\multicolumn{1}{c|}{} & -2$^{\circ}$ & 2$^{\circ}$ & \multicolumn{1}{c|}{-} & 51.52 ± 0.34 & 37.95 ± 3.70 \\
\multicolumn{1}{c|}{} & -3$^{\circ}$ & 3$^{\circ}$ & \multicolumn{1}{c|}{-} & 51.52 ± 0.34 & 39.04 ± 2.08 \\
\multicolumn{1}{c|}{} & -4$^{\circ}$ & 4$^{\circ}$ & \multicolumn{1}{c|}{-} & 54.19 ± 0.27 & 47.85 ± 0.42 \\
\multicolumn{1}{c|}{} & -5$^{\circ}$ & 5$^{\circ}$ & \multicolumn{1}{c|}{-} & 51.52 ± 0.34 & 38.88 ± 2.26 \\
\multicolumn{1}{c|}{} & -10$^{\circ}$ & 10$^{\circ}$ & \multicolumn{1}{c|}{-} & 51.52 ± 0.34 & 43.44 ± 0.44 \\ \midrule
\multicolumn{1}{c|}{\multirow{16}{*}{<Translation>}} & \multirow{2}{*}{-1px} & \multirow{2}{*}{1px} & \multicolumn{1}{c|}{True} & 62.56 ± 0.12 & \textbf{59.93} ± 0.91 \\
\multicolumn{1}{c|}{} &  &  & \multicolumn{1}{c|}{False} & \underline{62.73} ± 0.25 & 59.70 ± 0.95 \\
\multicolumn{1}{c|}{} & \multirow{2}{*}{-2px} & \multirow{2}{*}{2px} & \multicolumn{1}{c|}{True} & \textbf{62.78} ± 0.26 & \uwave{59.72} ± 0.92 \\
\multicolumn{1}{c|}{} &  &  & \multicolumn{1}{c|}{False} & \uwave{62.59} ± 0.25 & \underline{59.75} ± 1.02 \\
\multicolumn{1}{c|}{} & \multirow{2}{*}{-3px} & \multirow{2}{*}{3px} & \multicolumn{1}{c|}{True} & 62.30 ± 0.19 & 59.62 ± 1.19 \\
\multicolumn{1}{c|}{} &  &  & \multicolumn{1}{c|}{False} & 62.56 ± 0.14 & 59.71 ± 0.88 \\
\multicolumn{1}{c|}{} & \multirow{2}{*}{-4px} & \multirow{2}{*}{4px} & \multicolumn{1}{c|}{True} & 62.03 ± 0.20 & 59.46 ± 1.27 \\
\multicolumn{1}{c|}{} &  &  & \multicolumn{1}{c|}{False} & 61.73 ± 0.22 & 59.14 ± 1.39 \\
\multicolumn{1}{c|}{} & \multirow{2}{*}{-5px} & \multirow{2}{*}{5px} & \multicolumn{1}{c|}{True} & 61.51 ± 0.28 & 59.49 ± 1.31 \\
\multicolumn{1}{c|}{} &  &  & \multicolumn{1}{c|}{False} & 61.78 ± 0.11 & 58.82 ± 1.41 \\
\multicolumn{1}{c|}{} & \multirow{2}{*}{-10px} & \multirow{2}{*}{10px} & \multicolumn{1}{c|}{True} & 60.04 ± 0.15 & 58.33 ± 1.30 \\
\multicolumn{1}{c|}{} &  &  & \multicolumn{1}{c|}{False} & 59.23 ± 0.27 & 57.31 ± 1.37 \\
\multicolumn{1}{c|}{} & \multirow{2}{*}{-20px} & \multirow{2}{*}{20px} & \multicolumn{1}{c|}{True} & 56.53 ± 0.22 & 56.15 ± 1.09 \\
\multicolumn{1}{c|}{} &  &  & \multicolumn{1}{c|}{False} & 52.17 ± 0.16 & 52.14 ± 0.99 \\
\multicolumn{1}{c|}{} & \multirow{2}{*}{-30px} & \multirow{2}{*}{30px} & \multicolumn{1}{c|}{True} & 55.23 ± 0.36 & 54.30 ± 1.34 \\
\multicolumn{1}{c|}{} &  &  & \multicolumn{1}{c|}{False} & 51.52 ± 0.34 & 48.77 ± 0.88 \\ \bottomrule
\end{tabular}%
}
\vspace{-0.4cm}
\label{tab1}
\end{table}


\subsection{Detailed settings for existing state-of-the-art methods for comparison with \textbf{NBBOX}}
In this section, we briefly describe existing augmentation methods chosen for comparison with our approach.
Note that we implement all existing methods using their official repositories if possible.
Besides, we follow the recommended parameter settings if mentioned in their original papers.
However, if the original settings in the paper led to significantly lower performance than our baseline, we additionally conduct parameter tuning to ensure a fair comparison. 

\begin{table}[ht!]
\vspace{-0.3cm}
\centering
\caption{\small{\label{tab:table-name} Comparison of data augmentation methods: Bounding box vs image-level applications.}} \vspace{-0.2cm}

\resizebox{\columnwidth}{!}{%
\begin{tabular}{@{}cccccccccccccc@{}}
\toprule
 & RandRotate & RandShift & CutOut & AutoAug & RandAug & AugMix & MosaicMix & UniformAug & TrivialAug & YOCO & PixMix & InterAug & Ours \\ \midrule
BBox & O & O &  &  &  &  &  &  &  &  &  & O & O \\
Image & O & O & O & O & O & O & O & O & O & O & O & O &  \\ \bottomrule
\end{tabular}%
} \vspace{-0.2cm}
\label{tab2}
\end{table}

Table \ref{tab2} compares different data augmentation methods used in our experiments, specifically indicating their application at both the bounding box (BBox) level and the image level. 
    The methods include RandRotate \cite{lalitha2022review}, RandShift \cite{lalitha2022review}, CutOut \cite{devries2017improved}, AutoAug \cite{cubuk2019autoaugment}, RandAug \cite{cubuk2020randaugment}, AugMix \cite{hendrycks2020augmix}, MosaicMix \cite{takahashi2019data, wei2020amrnet}, UniformAug \cite{lingchen2020uniformaugment}, TrivialAug \cite{muller2021trivialaugment}, YOCO \cite{han2022you}, PixMix \cite{hendrycks2022pixmix}, InterAug \cite{devi2023improving}, and the proposed method, referred to as `Ours'. 
    Each method is marked with an `O' to indicate whether it is applied at the bounding box level or the image level.

    \vspace{0.1cm}
    Here, note that both RandRotate and RandShift involve transformations—rotation and translation, respectively—that require the bounding boxes to move alongside the image. 
        Therefore, these methods are marked with an `O' in both the BBox and Image rows.
    On the other hand, InterAug applies transformations to a portion of the image that contains the bounding box. As a result, both the BBox and Image levels are affected simultaneously. 
        Hence, InterAug is also marked with an `O' in both rows, indicating its application to both bounding boxes and images.

\vspace{0.1cm}
We briefly explain each augmentation method as follows:

\begin{description}
    \item [RandRotate.] 
Randomly rotates the image and bounding boxes with a probability of 0.5. 
    The rotation angle is randomly selected from a range of (-180, 180) degrees. 
    The rotation is applied while maintaining the aspect ratio of the bounding boxes. 
    During training, we randomly crop each image, and discard the cropped images (or patches) if they do not contain bounding boxes.

    \vspace{0.1cm}
    \item [RandShift.]
Randomly shifts the image and bounding boxes with a probability of 0.3. The shift is applied by randomly selecting a pixel offset in both the $x$ and $y$ directions within the range of (-64, 64) pixels for DIOR-R and (-128, 128) pixels for DOTA. 
    After shifting, any bounding box clipped to the image border with a width or height smaller than 1 pixel is discarded. 
    All the empty pixels due to the shift are filled with zero values. 

    \vspace{0.1cm}
    \item [CutOut.]
Square regions of pixels are randomly removed from the image, with the number of regions randomly selected from the interval [5, 10]. 
    This means that for each image, between 5 and 10 square areas are randomly chosen for removal. 
    For DIOR-R, the size of these square regions (or masks) is randomly chosen between 5$\times$5 pixels and 10$\times$10 pixels. 
    For DOTA, the mask size is larger, ranging between 10$\times$10 pixels and 20$\times$20 pixels. The locations of the dropped regions are uniformly sampled, and the dropped pixels are filled with black. 
    \item [AutoAug.]
AutoAug defines a search space where each policy consists of a series of sub-policies. 
    For each mini-batch, one sub-policy is randomly selected and applied during training. 
    Each sub-policy consists of two operations sampled from various transformations including shearing, rotation, or color jittering following the original paper. 
    Along with selecting the operations, the sub-policy also specifies the probability of applying each operation and the magnitude of the transformation. 
    These probabilities and magnitudes are learned and optimized to enhance model performance. 
 We follow the same search space as in the original paper.
    \vspace{0.1cm}
    \item [RandAug.]
RandAug consists of predefined augmentation operations selected from a reduced search space, where the strength or magnitude of each augmentation is easily controllable. 
    The key parameters in RandAug are the number of operations to apply and the magnitude of each transformation. 
    We used two augmentation operations per image with a fixed magnitude of 2. 
    This simplicity allows RandAug to be easily applied to various tasks while still offering effective data augmentation to enhance model performance. 
 We follow the same search space as in the original paper.
    \vspace{0.1cm}
    \item [AugMix.]
AugMix method blends several augmentation chains, each consisting of a sequence of transformations, to generate diverse image variations while preserving key features of the original image. 
    The key parameters in AugMix include severity, which controls the strength of the base augmentations applied to the image, and mixture width, which defines how many augmentation chains are combined to create the final image. 
    The chain depth determines the number of transformations in each chain. 
    Additionally, AugMix uses alpha, a parameter that controls the mixing proportions between augmentation chains, and supports a wide range of augmentation operations, including brightness, contrast, color, and sharpness adjustments. 
    We applied a `Bilinear' interpolation method to resize images for Augmix. 

    \vspace{0.1cm}
    \item [MosaicMix.]
Four images are combined into a single image (or mosaic) as output. 
    The mosaic is constructed by selecting a center point and arranging segments from each of the four images around the point. 
    To avoid out-of-memory issues, we set the mosaicked image size to 800$\times$800 for all datasets in this work.
    In most cases, padding is applied to maintain the final output size, with a default pad value of 114 (Gray color). 
    The augmentation is applied with a probability of 0.1, which means that only 10\% of the samples within each mini-batch are mosaicked. 

    \vspace{0.1cm}
    \item [UniformAug.]
UniformAug randomly selects multiple augmentation operations for each batch and applies them to the image with the same intensity. 
    Among the available augmentation options, multiple methods are randomly chosen and applied consistently across each batch. 
    The intensity (magnitude) of each augmentation is determined randomly but applied uniformly to all selected operations.
    We follow the same search space as in the original paper.
    
    \vspace{0.1cm}
    \item [TrivialAug.]
For each mini-batch, a single augmentation operation is randomly selected and applied.
    The magnitude of each augmentation is determined randomly within a predefined range.
     Despite its simplicity, it achieves performance improvements similar to more complex methods like AutoAug or RandAug, while reducing the complexity of the augmentation process.
  We follow the same search space as in the original paper.
  
    \vspace{0.1cm}
    \item [YOCO.]
YOCO is a simple way to augment image data that involves cutting an image into two equal-sized pieces, either in the height or width direction of the image, then performing separate data augmentations on each piece, and then combining the two pieces back together. 
    We randomly sample two methods from \{`ColorJitter', `RandomHorizontalFlip', `RandomVerticalFlip', `GaussianBlur', `RandomAffine', `RandomAdjustSharpness'\} and cropped the image vertically and horizontally for each iteration. 

    \vspace{0.1cm}
    \item [PixMix.]
PixMix involves blending the original image with another image from a fractal mixing set through multiple rounds of transformations, creating diverse image variations while maintaining the essential features of the original. 
    During each round, a random augmentation operation is applied to the original image or the mixing image, followed by a mixing operation such as additive or multiplicative blending. 
    It supports a variety of augmentation techniques, including rotation, solarization, and posterization, and generates highly varied training data to improve model robustness. 
    We applied a `Bilinear' interpolation method for resizing images in this process.

    \vspace{0.1cm}
    \item [InterAug.]
    Unlike traditional augmentation methods that apply transformations uniformly across an entire image or within predefined bounding boxes, InterAug dynamically determines the effective context around each object using bounding box annotations to guide augmentation. 
        In other words, it not only deforms the objects within bounding boxes but also adjusts the background and neighboring regions to maintain contextual consistency.
To further enhance the diversity of transformations, the authors combine InterAug with TrivialAug, a lightweight augmentation strategy that applies simple yet effective transformations like flipping, scaling, and color jittering.

\end{description}

\subsection{Analysis of experimental results}
Section III-C of the main manuscript presents a comparative analysis of our method against existing data augmentation schemes.
    In this section, we offer a detailed examination of the experimental results.
    In particular, we discuss \underline{why some methods achieve lower performance compared to the baseline without any data augmentation}.
\vspace{0.1cm}    
\begin{description}
\item [Severe data distortion.]
Some augmentation methods might apply transformations that excessively distort the data:
    \begin{itemize}
        \item (E.g. 1) RandShift shifts the image too much can result in objects being partially or entirely moved out of the frame.
        \item (E.g. 2) CutOut removes critical regions of the object can hinder the model from learning important features.
        \item (E.g. 3) in AugMix or RandAug, too highly challenging transformations early in training, causing the model to struggle in learning optimal features.
    \end{itemize}
    The core and shared problem among the above examples is that while appropriate data augmentation can help the model generalize better and improve its ability to grasp contextual cues, finding the optimal degree of augmentation is not straightforward.
    \textit{In contrast, our method uses predictable hyper-parameters, eliminating the burden of extensive tuning (almost parametric-free) while avoiding severe distortions to the data}. This ensures that the integrity of the contextual information is preserved, making the approach both efficient and reliable.
\vspace{0.1cm}
\item [Mismatch with data distribution.] 
If the augmented data significantly deviates from the distribution of the original dataset, the model may learn features that are not useful for the target data:
    \begin{itemize}
        \item (E.g. 1) MosaicMix combines portions of multiple images to create synthetic samples, which might differ too much from real-world data.
        \item (E.g. 2) InterAug applies transformations specifically to regions surrounding the objects (i.e. bounding box and its surroundings), instead of transforming an entire image, which does not naturally occur in real-world scenarios.
    \end{itemize}
    This discrepancy can cause the model to generalize poorly on actual test data.

\vspace{0.1cm}
\item [Lack of hyper-parameter tuning.]
The effectiveness of augmentation techniques depends heavily on hyperparameters like magnitude and probability.
    If these parameters are not tuned specifically for the dataset, the augmentation may degrade performance.
    Default parameters recommended in original papers might not be suitable for every dataset.
    Suboptimal parameters may either excessively distort the data or apply minimal transformations, preventing the augmentation from contributing to improved model performance.
    Note that, if the original settings introduced in other methods' papers led to significantly lower performance than our baseline, we additionally conduct parameter tuning to mitigate this issue.
    However, our exploration is not exhaustive due to resource and time constraints. 
    \textit{Data augmentation methods with optimal hyper-parameters may outperform our method and, furthermore, could be combined with our approach to achieve even greater performance improvements.}
     
\end{description}

\subsection{Time analysis of each data augmentation method used in this work}

In this section, Table \ref{tab3} indicates the time required per each training epoch on Faster R-CNN with ResNet-50 for each augmentation. 
To ensure fairness and objectivity, the experiments for each method were repeated three times with one training epoch, and the reported execution times are the averaged results of these three trials.
It is worth noting that our method is faster than other methods.
(E.g. for one epoch, ours vs \textcolor{black}{MosaicMix: 9.63min vs 21.24min on DIOR-R}). 

\begin{table}[ht!]
\vspace{-0.3cm}
\centering
\caption{\small{\label{tab:table-name} Time analysis of each transformation used in this work.}} \vspace{-0.2cm}

\resizebox{8cm}{!}{%
\begin{tabular}{@{}lcc@{}}
\toprule
\multicolumn{1}{c}{Detector: Faster R-CNN with ResNet-50} & DIOR-R & DOTA \\ \midrule
\multicolumn{1}{l|}{RandRotate } & \textcolor{white}{0}9.65min & 17.12min \\
\multicolumn{1}{l|}{RandShift }  & \textcolor{white}{0}9.66min & 16.37min \\
\multicolumn{1}{l|}{Cutout }  & \textcolor{white}{0}9.66min & 16.29min\\
\multicolumn{1}{l|}{AutoAug }  & \textcolor{white}{0}9.70min & 16.32min \\
\multicolumn{1}{l|}{RandAug }  & \textcolor{white}{0}9.64min & 16.34min \\
\multicolumn{1}{l|}{AugMix }  & \textcolor{white}{0}9.75min & 16.61min \\
\multicolumn{1}{l|}{MosaicMix } & 21.24min  & 34.48min \\
\multicolumn{1}{l|}{UniformAug } & \textcolor{white}{0}9.65min  & 16.34min \\
\multicolumn{1}{l|}{TrivialAug } & \textcolor{white}{0}9.65min  & 16.33min \\
\multicolumn{1}{l|}{YOCO } & \textcolor{white}{0}9.77min & 16.40min  \\
\multicolumn{1}{l|}{PixMix }  & \textcolor{white}{0}9.75min & 16.54min \\ 
\multicolumn{1}{l|}{InterAug } & \textcolor{white}{0}9.75min  & 17.10min \\
\multicolumn{1}{l|}{\textbf{Ours (Scaling + Rotation + Translation)}}  & \textbf{\textcolor{white}{0}9.63min} & \textbf{16.28min} \\
\bottomrule
\end{tabular}%
} \vspace{-0.3cm}
\label{tab3}
\end{table}

\end{document}